\documentclass[sigconf]{acmart}

\usepackage{algorithm}
\usepackage{algorithmic}
\usepackage{multirow}
\usepackage{graphics}
\usepackage{stfloats}
\usepackage{enumitem}
\usepackage{subfig}
\usepackage{float}
\usepackage{hyperref}

\newtheorem{lemma}{Lemma}

\makeatother

\newcommand{\gnn}{g_\theta}
\newcommand{\influence}{\mathrm{F}_{\gnn}}

\newcommand{\model}{{NORA}}

\AtBeginDocument{%
}

\setcopyright{acmcopyright}
\copyrightyear{2024}
\acmYear{2024}
\setcopyright{rightsretained}
\acmConference[WWW '24]{Proceedings of the ACM Web Conference 2024}{May 13--17, 2024}{Singapore, Singapore}
\acmBooktitle{Proceedings of the ACM Web Conference 2024 (WWW '24), May 13--17, 2024, Singapore, Singapore}
\acmDOI{10.1145/3589334.3645389}
\acmISBN{979-8-4007-0171-9/24/05}

\makeatletter
\gdef\@copyrightpermission{
\begin{minipage}{0.3\columnwidth}
\href{https://creativecommons.org/licenses/by/4.0/}{\includegraphics[width=0.90\textwidth]{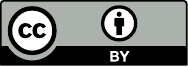}}
\end{minipage}\hfill
\begin{minipage}{0.7\columnwidth}
\href{https://creativecommons.org/licenses/by/4.0/}{This work is licensed under a Creative Commons Attribution International 4.0 License.}
\end{minipage}
\vspace{5pt}
}
\makeatother

\acmSubmissionID{434}

\begin{document}

\title{Fast Inference of Removal-Based Node Influence}


\author{Weikai Li}
\email{weikaili@cs.ucla.edu}
\affiliation{
  \institution{University of California, Los Angeles}
  \state{California}
  \country{USA}
}

\author{Zhiping Xiao}
\email{patricia.xiao@gmail.com}
\affiliation{
  \institution{University of California, Los Angeles}
  \state{California}
  \country{USA}
}

\author{Xiao Luo}
\email{xiaoluo@cs.ucla.edu}
\affiliation{
  \institution{University of California, Los Angeles}
  \state{California}
  \country{USA}
}

\author{Yizhou Sun}
\email{yzsun@cs.ucla.edu}
\affiliation{
  \institution{University of California, Los Angeles}
  \state{California}
  \country{USA}
}




\begin{abstract}

Graph neural networks (GNNs) are widely utilized to capture the information spreading patterns in graphs. While remarkable performance has been achieved, there is a new trending topic of evaluating node influence. We propose a new method of evaluating node influence, which measures the prediction change of a trained GNN model caused by removing a node. A real-world application is, ``In the task of predicting Twitter accounts' polarity, had a particular account been removed, how would others' polarity change?''. We use the GNN as a surrogate model whose prediction could simulate the change of nodes or edges caused by node removal. Our target is to obtain the influence score for every node, and a straightforward way is to alternately remove every node and apply the trained GNN on the modified graph to generate new predictions. It is reliable but time-consuming, so we need an efficient method. The related lines of work, such as graph adversarial attack and counterfactual explanation, cannot directly satisfy our needs, since their problem settings are different. We propose an efficient, intuitive, and effective method, \textbf{NO}de-\textbf{R}emoval-based f\textbf{A}st GNN inference (\textbf{\model}), which uses the gradient information to approximate the node-removal influence. It only costs one forward propagation and one backpropagation to approximate the influence score for all nodes. Extensive experiments on six datasets and six GNN models verify the effectiveness of {\model}. Our code is available at https://github.com/weikai-li/NORA.git.
\end{abstract}

\ccsdesc[500]{Information systems~Web mining}
\ccsdesc[500]{Computing methodologies~Neural networks}

\keywords{Graph Neural Network, Web Mining, Node Influence Evaluation}


\maketitle

\section{Introduction}\label{sec::intro}

Measuring node influence in a graph and identifying influential nodes are important to various applications, such as advertising~\cite{FirstIM,IM_approximation2,viral_marketing1}, online news dissemination~\cite{news_dissemination1,news_dissemination2}, finding bottlenecks in an infrastructure network to improve robustness~\cite{real_removal_example2,real_removal_example3}, vaccination on prioritized groups of people to break down virus spreading~\cite{node_removal_review,disease_example1,disease_example2}, etc.
This topic has attracted many studies. ``Influence maximization'' problem aims to identify influential nodes whose triggered influence spreading range is maximized~\cite{attribute_IM1,attribute_IM2,attribute_IM4,attribute_IM8,IM_GNN3,IM_GNN4,IM_GNN5,IM_GNN6}. ``Network dismantling'' problem studies the influence of node removal on network connectivity\cite{node_removal3,node_removal4,node_removal5,node_removal_neural1,node_removal_neural2,node_removal_transfer}.

\begin{figure}[tbp]
\centering
\includegraphics[width=0.95\linewidth]{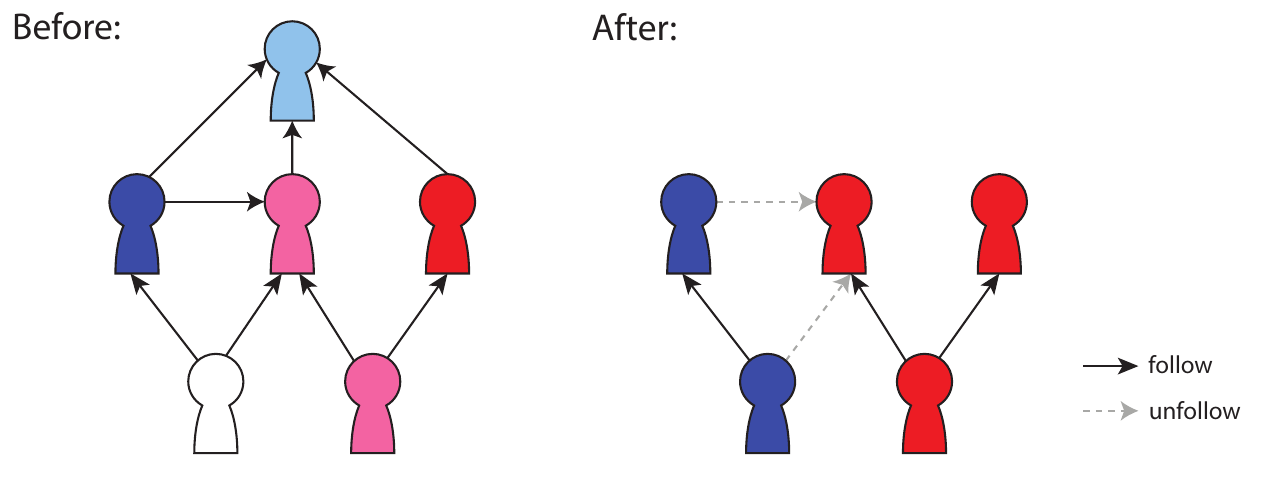}
\caption{An example of the task-specific influence of node removal in social networks. Red versus Blue represents two different opinions, and color shades represent the degree of opinion. When the top blue node is removed, the two pink nodes might hear less voice from the blue nodes and become red. The two left nodes might no longer follow the middle node, and the left white node might become blue. These are the influence of removing the top blue node.}
\label{fig:node_removal_example}
\end{figure}

They define the node influence based on the graph structure (e.g., connectivity) or a designed information propagation model (e.g., the susceptible-infected-removed model). However, these definitions are not flexible enough to capture the node influence from different aspects. For example, we might want to identify the biggest political influencers on Twitter. In this case, we want to calculate the influence score of a Twitter account based on how much it would affect other users' political polarity had it been removed. In another scenario, we might want to identify the biggest fashion influencers on Twitter, and we want to calculate the node influence based on how much it would affect other users' fashion categories had it been removed. Instead of adopting any fixed node influence definition, we thus focus on task-specific node influence score calculation based on node removal. Figure~\ref{fig:node_removal_example} provides an example of the task-specific node influence. 

\begin{figure}[h]
\centering
\includegraphics[width=0.48\textwidth]{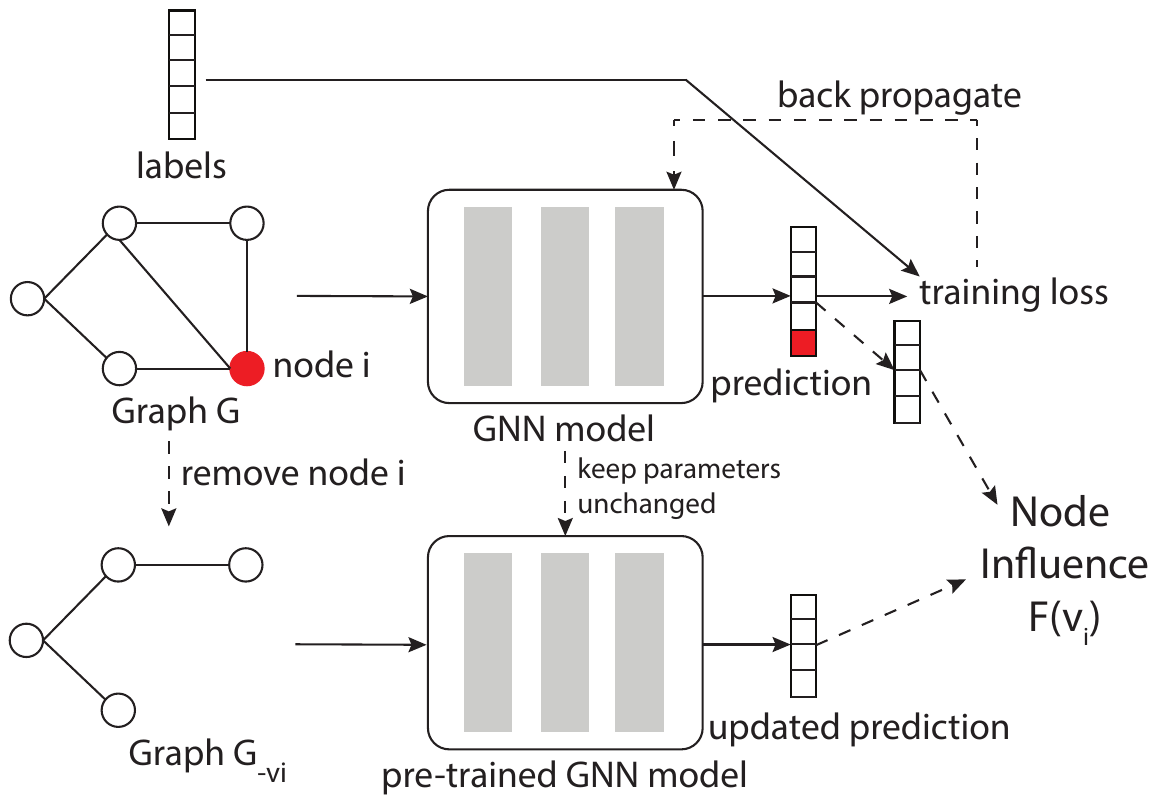}
\caption{Our schema of calculating node influence. The GNN model is trained on the original graph. We remove a node and apply the trained GNN to the modified graph. We calculate the total variation distance between the original predictions and new predictions as the influence of node removal.}
\label{fig:influence_definition}
\vspace{-0.3cm}
\end{figure}

Graph neural networks (GNNs) are among the most powerful graph learning tools. We use GNNs as a surrogate to capture the underlying mechanism of how the graph structure affects node behaviors. 
In the ideal case, we should train a new GNN based on the graph where the target node is removed and other node/edge labels could also be different. Unfortunately, this graph only exists in a parallel world and is not available for training GNNs. We have two options to solve this issue. First, we can assume the model does not change significantly, so we can use the model trained on the original graph. Second, we can assume the labels of other nodes/edges do not change, so we can re-train the GNN on the new graph where the target node is removed and the labels do not change. We choose the first option, because the underlying patterns of message spreading learned by the GNN should be relatively stable. After removing a node, we use the new predictions of the trained GNN on the modified graph to simulate the new labels in the parallel world.
We calculate the influence of node removal as the total variation distance between the original predictions and new predictions, as illustrated in Figure~\ref{fig:influence_definition}.

Our target is to calculate the influence score for every node. The most straightforward way is to alternately remove each node from the original graph and use the trained GNN to do prediction. However, it is very time consuming, so we demand an efficient method. Evaluating the change of GNN predictions caused by the change of input has been studied in graph adversarial attack and graph counterfactual explanation. Graph adversarial attack aims to maximally undermine GNN performance or change GNN prediction by perturbations to the input graph, which mainly include modifying node features~\cite{NETTACK,NETTACK_long,InfMax}, injecting nodes~\cite{first_injection_attack,second_injection_attack,TDGIA,HAO,TUA,GA2C,G-NIA}, or edge perturbation~\cite{Meta,PGD,CLGA,edge_attack}. However, to the best of our knowledge, none of the adversarial attack methods utilizes node removal, since it is not common in the attack scenario.

Graph counterfactual explanation aims to explain the GNN's prediction of a target node/edge/graph by finding the minimum perturbation on the input graph that can change the prediction of the target~\cite{counterfactual}. Some methods utilize node removal~\cite{CF2,counterfact_node_edge,LARA,CLEAR,XGNN,SubgraphX,GCFExplainer}, but they can not directly satisfy our needs. First, our problem setting is different. We evaluate the influence of removing one node on other nodes/edges, while the explanation methods focus on the influence of removing several nodes on one target node/edge/graph.
Second, the existing strategies can not easily scale up to handle large graphs when the goal is to predict the influence score for every node.
We briefly summarize the difference in problem settings between related lines of research and this paper in Table~\ref{tab:related_work}. There are three important aspects in our problem setting: task-specific, influence of node removal, and global influence (influence on all nodes/edges). As shown in the table, the existing lines of work generally do not simultaneously have these three aspects.

\begin{table}[t]
    \centering
    \small
    \caption{Differences in problem settings.}
    \label{tab:related_work}
    \scalebox{0.93}{
    \begin{tabular}{lrrr}
    \toprule
         & Task-specific & Removal & Global influence \\  
    \midrule
     Influence maximization &  &  & \checkmark \\
     Network dismantling &  & \checkmark & \checkmark \\
     Adversarial attack & \checkmark &  & \checkmark \\
     Counterfactual explanation & \checkmark & \checkmark & \\
     This paper & \checkmark & \checkmark & \checkmark \\
    \bottomrule
    \end{tabular}
    }
\end{table}

To efficiently calculate the node influence score, we propose an intuitive, effective, and efficient method, \textbf{NO}de-\textbf{R}emoval-based f\textbf{A}st GNN Inference (\textbf{\model}). We use the first-order derivatives to approximate the influence. It is model-agnostic and can be easily adapted to any common GNN model based on the message-passing framework. It only needs one forward propagation and one backpropagation to approximate the removal influence for all nodes. It takes up to 41 hours to generate the real node influence in our experiments, while {\model} uses less than a minute if we do not include the time of generating the validation-set labels. Although simple and intuitive, {\model} works well in our experiments. We modify and adapt two approaches in graph counterfactual explanation as baselines, and {\model} outperforms them in the experiments. To sum up, this paper makes the following contributions:

\begin{itemize}[leftmargin=*]
    \item
    \textit{New Problem}. We propose a novel perspective of evaluating task-specific node influence based on node removal by GNN.
    
    \item
    \textit{Methodology}. We propose an efficient, intuitive, and effective algorithm, {\model}, 
    to approximate the node influence for all nodes.
    
    \item
    \textit{Effectiveness.} Experimental results on six datasets and six GNN models demonstrate that {\model} outperforms the baselines.
\end{itemize}

\section{Related Work}

\subsection{Graph Adversarial Attack}
Graph adversarial attack aims to maximally undermine GNN performance or change GNN predictions by imposing a small perturbation. Zügner et al.~\cite{NETTACK,NETTACK_long} started the race of graph adversarial attacks. Pioneering works mainly focused on modifying node features~\cite{NETTACK,NETTACK_long,InfMax} and perturbing edges~\cite{Meta,PGD,CLGA,edge_attack}. Some recent works~\cite{first_injection_attack,second_injection_attack,TDGIA,GA2C,G2A2C,HAO,TUA,G-NIA} study the node injection attack, which injects new nodes into a graph and connects them with some existing nodes. Chen et al.~\cite{HAO} prove that the node injection attack can theoretically cause more damage than the graph modification attack with less or equal modification budget. G-NIA model~\cite{G-NIA} sets a strong limitation that the attacker can only inject one node with one edge, and it achieves more than 90\% successful rate in the single-target attack on Reddit and ogbn-products datasets. They demonstrate the strong potential of altering nodes' existence, which is inspiring to our research. To the best of our knowledge, none of the adversarial attack methods considers node removal, since it is not common in attacking applications.

\subsection{Graph Counterfactual Explanation} 
Graph counterfactual explanation aims to explain why a GNN model gives a particular result of a target node/edge/graph.
Unlike the factual explanation that explains by associating the prediction with critical nodes or edges, the counterfactual explanation tries to find the minimum perturbation on the input graph that can change the prediction of the target. Some methods~\cite{GNNExplainer,CF-GNNExplainer,RCExplainer,PGExplainer} are purely based on edge removal; some methods utilize both node removal and edge removal, and the methods include optimizing mask matrices~\cite{CF2,counterfact_node_edge}, predicting node influence~\cite{LARA}, applying graph generation models~\cite{CLEAR,XGNN}, or searching for an optimal neighbor graph~\cite{SubgraphX,GCFExplainer}. As analyzed in Section~\ref{sec::intro} and shown in Table~\ref{tab:related_work}, our problem setting differs from existing works, so these methods are not directly applicable. We modify and adapt two widely used graph counterfactual explanation methods as the baseline methods. The first baseline is inspired by optimizing a mask matrix~\cite{CF-GNNExplainer,GNNExplainer,CF2,RCExplainer,counterfact_node_edge}. The mask matrix indicates edge existence, and it is multiplied by the adjacency matrix before the message passing. Its elements are within $[0, 1]$, and the mask matrix is optimized to maximize the difference between original predictions and new predictions. We adapt it to node removal by using the node mask vector. Our second baseline is inspired by LARA~\cite{LARA}, which trains a GCN model to predict node influence on the explanation target. The parameter size does not grow with the input graph size, so it is more scalable compared to previous methods.

\section{Problem Definition}

\subsection{Notations}

A graph $G=(V,E)$ consists of nodes $\boldsymbol{V}=\{v_1, v_2, ..., v_N\}$ and edges $E = \{ e_{ij} | j \in \mathcal{N}(i) \}$, where $\mathcal{N}(i)$ denotes the neighbor nodes of $v_i$ and $e_{ij}$ denotes the edge from $v_i$ to $v_j$. We use $\hat{\mathcal{N}}(i)$ to denote $\mathcal{N}(i) \cup \{v_i\}$. $\boldsymbol{A}$ denotes the adjacency matrix. Node $v_i$ is associated with feature vector $\boldsymbol{x_i} \in \mathbb{R}^d$, and a label $y_i \in \mathbb{R}$ if the node classification task is applicable. We denote the degree of $v_i$ as $d_i = |\mathcal{N}(i)|$. When we remove node $v_r \in V$, we also remove all edges connected with $v_r$ from graph $G$, and we denote the modified graph as $G_{-v_r}$. $\gnn$ denotes a trained GNN model. We use $v_r$ to denote the target removing node and $\influence(v_r)$ to denote the influence of removing $v_r$.

Graph neural networks (GNNs) generally follow the message-passing framework~\cite{MPNN}. A GNN model consists of multiple graph convolutional layers. In a typical graph convolutional layer, a node updates its representation by aggregating its neighbor nodes' representations:
\begin{equation}
    \boldsymbol{h_i^{(l)}} = U_l(\boldsymbol{h_i^{(l-1)}}, \mathrm{AGG}_l(\sum_{j \in \mathcal{\hat{N}}(i)} \mathrm{MSG}_l(\boldsymbol{h_j^{(l-1)}}, \boldsymbol{h_i^{(l-1)}}))),
\label{equa:mpnn_framework}
\end{equation}
where $\boldsymbol{h_i^{(l)}}$ denotes the representation of $v_i$ after the $l$-th layer ($l \in {1, 2, \dots}$), and $\boldsymbol{h_i^{(0)}}$ is the input feature $\boldsymbol{x_i}$. $\mathrm{MSG}_l$ is the message function, $\mathrm{AGG}_l$ is the aggregation function, and $U_l$ is the update function.

\subsection{Problem Definition}

To evaluate the influence of node removal, we use GNN models as a surrogate to predict the change of nodes/edges caused by removing the target node.
As shown in Figure~\ref{fig:influence_definition}, we measure the influence by the total variation distance between the original and updated predicted probability distribution, and we use the $\ell_1$-norm of the difference, which can equally capture the prediction change for every class.

\noindent\textbf{Definition 1. (Node Influence in Node Classification Task)} Given a node classification model $\gnn$ trained on graph $G$, we denote its predicted class probability of node $v_i$ as $\boldsymbol{\gnn(G)_i} \in \mathbb{R}^c$ ($c$ is the number of classes), the influence of node $v_r$ is calculated as:
\begin{equation}
    \influence(v_r) = \sum_{i=1,i\neq r}^{N} ||\boldsymbol{\gnn(G)_i} - \boldsymbol{\gnn(G_{-v_r})_i}||_1\,,
\label{equa:influence_definition_node}
\end{equation}

\noindent\textbf{Definition 2. (Node Influence in Link Prediction Task)} Given a link prediction model $\gnn$ trained on graph $G$, we denote its predicted probability of edge $e_{ij}$ as $f_e(\gnn(G)_{e_{ij}}) \in \mathbb{R}$, where $f_e(\cdot)$ is the optional layers that transform $\gnn$'s represetations to the predicted edge probability. We use $D_e$ to denote the whole link prediction set, and $D_r$ to denote edges that connect to $v_r$. The influence of removing node $v_r$ is calculated as:
\begin{equation}
    \influence(v_r) = \sum_{e_{ij} \in D_e - D_r} |f_e(\gnn(G)_{e_{ij}}) - f_e(\gnn(G_{-v_r})_{e_{ij}})|\,,
\label{equa:influence_definition_edge}
\end{equation}

\section{Methods}

\begin{figure}[t]
\centering
\includegraphics[width=\linewidth]{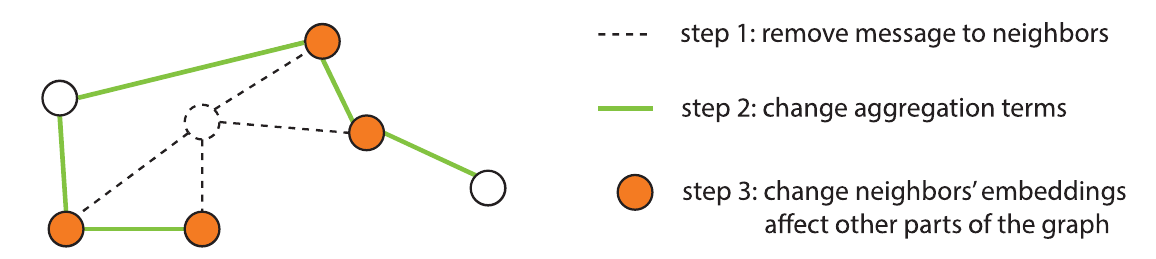}
\caption{Three kinds of influence of node removal: the disappearance of its node embedding; the change of its nearby nodes' aggregation terms; and the spread-out influence to multi-hop neighbors.}
\label{fig:model_algorithm_steps}
\end{figure}

Our target is to calculate the defined influence score for every node on a graph. The biggest challenge is efficiency, and we want to avoid traversing all the nodes. We propose an intuitive yet effective method, \textbf{NO}de-\textbf{R}emoval-based f\textbf{A}st GNN inference (\textbf{\model}). In general, we approximate the node influence by analyzing the calculation formula and decomposing it into three parts, which correspond to three kinds of influence of the node removal. Then, we use gradient information and some heuristics to approximate it. Figure~\ref{fig:model_algorithm_steps} illustrates the three kinds of influence. We mainly introduce our method in the node classification task, and after that, we will explain how to generalize it to the link prediction task.

\subsection{Influence Score Calculation Decomposition}


We cannot directly use the first-order derivatives to approximate node influence based on the definition in Equation~\ref{equa:influence_definition_node}, since there is a $\ell_1$-norm inside the summation. Intuitively, removing a node usually causes consistent change to the class of other nodes, e.g., raising/lowering the probability of some classes for all nodes. Thus, we approximate by moving the $\ell_1$-norm out of the summation. We denote the number of GNN layers as $L$, and $\boldsymbol{h_i^{(L)}} \in R^c$ is the predicted class probability of node $v_i$ ($c$ is the number of classes). We denote as $\boldsymbol{f_r} = \sum_{i=1, i \neq r}^N \boldsymbol{h_i^{(L)}}$ the sum of all nodes' predictions except for node $v_r$, and we denote as $\boldsymbol{\delta f_r}$ the change of $\boldsymbol{f_r}$ caused by removing node $v_r$.

\begin{lemma}
If removing $v_r$ consistently changes the class distributions of other nodes, its influence defined in Equation~\ref{equa:influence_definition_node} is equal to:
\begin{small}
\begin{gather}
    ||\sum_{i\neq r} (\boldsymbol{\gnn(G)_i} - (\boldsymbol{\gnn(G_{-v_r})_i})||_1
    = ||\boldsymbol{\delta f_r}||_1
    = ||\sum_{i \neq r} \boldsymbol{\frac{\partial f_r}{\partial h_i^{(L)}}} \boldsymbol{\delta h_i^{(L)}}||_1,
\label{equa:influence_definition_alter}
\end{gather}
\end{small}
\end{lemma}
where $\boldsymbol{\delta h_i^{(L)}}$ is the change of $\boldsymbol{h_i^{(L)}}$ caused by removing $v_r$. The formula above is strictly equal because $\boldsymbol{\frac{\partial f_r}{\partial h_i^{(L)}}} = \boldsymbol{1}$. We write it in this form because we want to keep a uniform form with later formulas. We can extend this form from the last layer to the previous layers. Here we analyze how to extend it from the $L$-th layer to the $(L-1)$-th layer.
Equation~\ref{equa:mpnn_framework} illustrates the framework of a message-passing GNN layer. We consider a typical parameterization of it:
\begin{equation}
    \boldsymbol{h_i^{(l)}} = \sigma \big( \boldsymbol{W_u^{(l)}} (\boldsymbol{W_s^{(l)}} \boldsymbol{h_i^{(l-1)}} + \sum_{j \in \hat{\mathcal{N}}(i)} \alpha_{ji}^{(l)} \boldsymbol{W_m^{(l)}} \boldsymbol{h_j^{(l-1)}}) \big),
\label{equa:gradient1}
\end{equation}
where $\sigma$ denotes the activation function, $\boldsymbol{W_u^{(l)}}$, $\boldsymbol{W_s^{(l)}}$, and $\boldsymbol{W_m^{(l)}}$ are model parameters. $\alpha_{ji}^{(l)}$ is the normalization term.
The model parameters are fixed during inference.
Therefore, we can approximate $\boldsymbol{\delta h_i^{(L)}}$ by the first-order derivatives as:
\begin{gather}
    \boldsymbol{\delta h_i^{(L)}} \approx - I(v_r \in \mathcal{N}(i)) \boldsymbol{\frac{\partial h_i^{(L)}}{\partial h_r^{(L-1)}}}  \boldsymbol{h_r^{(L-1)}} \nonumber \\
    + \sum_{j \in \hat{\mathcal{N}}(i), j \neq r} (\boldsymbol{\frac{\partial h_i^{(L)}}{\partial \alpha_{ji}^{(L)}}} \delta\alpha_{ji}^{(L)} + \boldsymbol{\frac{\partial h_i^{(L)}}{\partial h_j^{(L-1)}}} \boldsymbol{\delta h_j^{(L-1)}}),
\label{equa:gradient3}
\end{gather}
where $I(.)$ is the indicator function. By incorporating the definition of $\boldsymbol{\delta f_r}$, we can derive the following formula.

\begin{lemma}
We can approximate $\boldsymbol{\delta f_r}$ for the GNN model described in Equation~\ref{equa:gradient1} using the first-order derivatives as:
\begin{small}
\begin{gather}
    \boldsymbol{\delta f_r} \approx - T_1 + T_2 + T_3 = - \sum_{i \in \mathcal{N}(r)} \boldsymbol{\frac{\partial f_r}{\partial h_i^{(L)}}}  \boldsymbol{\frac{\partial h_i^{(L)}}{\partial h_r^{(L-1)}}}  \boldsymbol{h_r^{(L-1)}} + \sum_{i \neq r}\sum_{j \in \hat{\mathcal{N}}(i), j \neq r} \nonumber \\
    (\boldsymbol{\frac{\partial f_r}{\partial h_i^{(L)}}}  \boldsymbol{\frac{\partial h_i^{(L)}}{\partial \alpha_{ji}^{(L)}}}  \delta \alpha_{ji}^{(L)}) + 
    \sum_{i \neq r}\sum_{j \in \hat{\mathcal{N}}(r), j \neq r} (\boldsymbol{\frac{\partial f_r}{\partial h_i^{(L)}}}  \boldsymbol{\frac{\partial h_i^{(L)}}{\partial h_j^{(L-1)}}}  \boldsymbol{\delta h_j^{(L-1)}}).
\label{equa:gradient4}
\end{gather}
\end{small}
\end{lemma}
The calculation contains three terms. $T_1$ measures the disappearance of $v_r$'s latent representation as a message to its neighbor nodes; $T_2$ measures the change of its neighbors' normalization term; $T_3$ measures the change of its neighbors' latent representations. The three terms correspond to the three kinds of influence in Figure~\ref{fig:model_algorithm_steps}.

\subsection{Approximation of Each Decomposed Term}

\noindent \textbf{$T_1$: Disappearance of the message to neighbor nodes.}
On the computation graph, $\boldsymbol{h_r^{(L-1)}}$ can connect to later layers either by $\boldsymbol{h_r^{(L)}}$ or by $\boldsymbol{h_i^{(L)}}, i \in \mathcal{N}(r)$. Therefore, by applying the chain rule, the term $T_1$ is equal to:
\begin{equation}
    \label{equa:grad_3_1}
    \boldsymbol{\frac{\partial f_r}{\partial h_r^{(L-1)}}} \boldsymbol{h_r^{(L-1)}} - \boldsymbol{\frac{\partial f_r}{\partial h_r^{(L)}}} \boldsymbol{\frac{\partial h_r^{(L)}}{\partial h_r^{(L-1)}}} \boldsymbol{h_r^{(L-1)}}.
\end{equation}
Although $\boldsymbol{h_r^{(L)}}$ is not related to $\boldsymbol{f_r}$ and $\boldsymbol{\frac{\partial f_r}{\partial h_r^{(L)}}} = \boldsymbol{0}$, we still write it in this way as a general form which can be later applied to previous layers, since $\boldsymbol{h_r^{(L-1)}}$, $\boldsymbol{h_r^{(L-2)}}, etc.$ is related to $\boldsymbol{f_r}$. Equation~\ref{equa:grad_3_1} consists of two parts. The form of the first part is more convenient to handle, so we want to eliminate the second part. We do this by approximating the ratio of the second part to the first part. Here we make a rough assumption that every node is functionally and structurally equal, which means they have the same degree, the same representation, and the same gradient. We denote the gradient coming from a neighbor, $\boldsymbol{\frac{\partial f_r}{\partial h_i^{(L)}}} \boldsymbol{\frac{\partial h_i^{(L)}}{\partial h_r^{(L-1)}}}$, as $\boldsymbol{g}$, and the gradient coming from the higher-layer representation of a node itself, $\boldsymbol{\frac{\partial f_r}{\partial h_r^{(L)}}} \boldsymbol{\frac{\partial h_r^{(L)}}{\partial h_r^{(L-1)}}}$, as $\beta \boldsymbol{g}$. $\beta \in \mathbb{R}$ is usually larger than one due to self-loop and residual connection. Then, the ratio of the second part versus the first part in Equation~\ref{equa:grad_3_1} can be approximated as $\frac{\beta}{d_r+\beta}$, where $d_r$ is the degree of node $v_r$. Then we derive the following lemma.

\begin{lemma}
If every node in the graph is structurally and functionally equal, we can approximate the term $T_1$ in Equation~\ref{equa:gradient4} as:
\begin{gather}
    \label{equa:grad_3_2}
    T_1 \approx \frac{d_r}{d_r+\beta} \boldsymbol{\frac{\partial f_r}{\partial h_r^{(L-1)}}} \boldsymbol{h_r^{(L-1)}}.
\end{gather}
\end{lemma}
For computation convenience, we approximate $\boldsymbol{\frac{\partial f_r}{\partial h_r^{(L-1)}} h_r^{(L-1)}}$ by a scalar. Based on experiments, we find that an effective way to approximate $\boldsymbol{\frac{\partial f_r}{\partial h_r^{(L-1)}} h_r^{(L-1)}}$ is calculating $||(\boldsymbol{f_r \frac{\partial f_r}{\partial h_r^{(L-1)}}) \circ h_r^{(L-1)}}||_p$, where $\circ$ is element-wise product and $||.||_p$ is the $\ell_p$-norm. Here we multiply the original formula by $\boldsymbol{f_r}$ so that we increase the focus on the classes with high predicted probabilities. Here $\boldsymbol{f_r} \in \mathbb{R}^{c}$, $\boldsymbol{\frac{\partial f_r}{\partial h_r^{(L-1)}}} \in \mathbb{R}^{cxd}$ is the Jacobian matrix, and $\boldsymbol{h_r^{(L-1)}} \in \mathbb{R}^d$. p is a hyper-parameter, and in most cases, we set it to one.

\noindent \textbf{$T_2$: Change of aggregation terms.}
There are two challenges in approximating the term $T_2$. First, calculating $\boldsymbol{\frac{\partial f_r}{\partial h_i^{(L)}} \frac{\partial h_i^{(L)}}{\partial \alpha_{ji}^{(L)}}}$ might consume too much space on large dense graphs with many edges. Second, the aggregation term differs significantly in different GNNs. For example, GCN~\cite{GCN} and GraphSAGE~\cite{graphsage} use node degree to compute it, while some models use the attention mechanism, such as GAT~\cite{GAT}. Therefore, it is difficult to derive a generally effective approximation.

After several attempts, we find that a widely effective way is to ignore $\boldsymbol{\frac{\partial f_r}{\partial h_i^{(L)}} \frac{\partial h_i^{(L)}}{\partial \alpha_{ji}^{(L)}}}$ and approximate $\delta \alpha_{ji}^{(L)}$ only by structure. We combine the aggregation term of GCN~\cite{GCN} and GraphSAGE~\cite{graphsage}. The aggregation of GCN is $\alpha_{ji}^{(L)} = 1/\sqrt{d_i d_j}$, and that of GraphSAGE (with mean aggregation) is $\alpha_{ji}^{(L)} = 1/d_i$. If $v_i$ is not $v_r$'s neighbor, then we assume the aggregation term does not change. If $v_i$ is $v_r$'s neighbor, we approximate $\delta \alpha_{ji}^{(L)}$ by $\hat{\delta} \alpha_{ji}$ as:
\begin{gather}
    \hat{\delta} \alpha_{ji} = [k_1 ( \frac{1}{\sqrt{d_i-1}} - \frac{1}{\sqrt{d_i}}) + (1 - k_1) (\frac{1}{d_i-1} - \frac{1}{d_i})] \nonumber \\
     [k_2 \frac{1}{\sqrt{d_j}} + k_2' \frac{1}{d_j} + (1 - k_2 - k_2')],
\label{equa:grad_2_1}
\end{gather}
where $k_1$, $k_2$, and $k_2'$ are hyper-parameters within [0,1]. There exist hyper-parameters that make $\hat{\delta} \alpha_{ji}$ satisfy GCN or GraphSAGE. Based on $\hat{\delta} \alpha_{ji}$, we approximate the second term as:
\begin{equation}
    T_2 \approx k_3 \cdot \delta Topo_r,\quad \delta Topo_r= \sum_{i \in \mathcal{N}(r)} \sum_{j \in \mathcal{N}(i)} \hat{\delta} \alpha_{ji},
    \label{equa:topo}
\end{equation}

where $k_3$ is a hyper-parameter.

\noindent\textbf{$T_3$: Change of hidden representations of other nodes.} When we analyze the term $T_3$, on the computation graph, $\boldsymbol{h_j^{(L-1)}}$ can reach $\boldsymbol{f_r}$ by either $\boldsymbol{h_r^{(L)}}$ or $\boldsymbol{h_i^{(L)}}, i \neq r$. Based on the chain rule, we can simplify $T_3$ in Equation~\ref{equa:gradient4} to Equation~\ref{equa:gradient_1_1}, which can be further transformed into Equation~\ref{equa:gradient_1_2}.
\begin{gather}
    \sum_{j \neq r} (\boldsymbol{\frac{\partial f_r}{\partial h_j^{(L-1)}}}  - \boldsymbol{\frac{\partial f_r}{\partial h_r^{(L)}}} \boldsymbol{\frac{\partial h_r^{(L)}}{\partial h_j^{(L-1)}} )} \boldsymbol{\delta h_j^{(L-1)}} \label{equa:gradient_1_1} \\
    = \sum_{j \neq r} \boldsymbol{\frac{\partial f_r}{\partial h_j^{(L-1)}}} \boldsymbol{\delta h_j^{(L-1)}} - \sum_{j \in \mathcal{N}(r)} \boldsymbol{\frac{\partial f_r}{\partial h_r^{(L)}} \frac{\partial h_r^{(L)}}{\partial h_j^{(L-1)}} \delta h_j^{(L-1)}}.
    \label{equa:gradient_1_2}
\end{gather}
Although $\boldsymbol{h_r^{(L)}}$ is not related to $\boldsymbol{f_r}$ and $\boldsymbol{\frac{\partial f_r}{\partial h_r^{(L)}}} = \boldsymbol{0}$, we still write it in this way as a general form which can be later applied to previous layers, since $\boldsymbol{h_r^{(L-1)}}$, $\boldsymbol{h_r^{(L-2)}}, etc.$ is related to $\boldsymbol{f_r}$. Similar to the simplification process of $T_1$, here we also arrive at a formula with two parts. The form of the first part is more convenient to handle, and it takes the same form as Equation~\ref{equa:influence_definition_alter}, so we want to eliminate the second part. We make the same rough assumption as approximating $T_1$ that every node is functionally and structurally equal. We denote the average node degree as $d$ and $\boldsymbol{\delta h_j^{(L-1)}}$ as $\boldsymbol{\delta h}$. Using the same notations of $\boldsymbol{g}$ and $\beta \boldsymbol{g}$ as approximating $T_1$, we approximate $\boldsymbol{\frac{\partial f_r}{\partial h_j^{(L-1)}}}$ as $(d+\beta)\boldsymbol{g}$, and we approximate the first part of Equation~\ref{equa:gradient_1_2} as $(N-1)(d+\beta)\boldsymbol{g\delta h}$. We approximate the second part of Equation~\ref{equa:gradient_1_2} as $d_r \boldsymbol{g\delta h}$. Then by rewriting the enumeration variable $j$ as $i$, we derive the following lemma.
\begin{lemma}
If every node in the graph is structurally and functionally equal, we can approximate the term $T_3$ in Equation~\ref{equa:gradient4} as:
\begin{gather}
    T_3 \approx T_3' = \big( \sum_{i \neq r} \boldsymbol{\frac{\partial f_r}{\partial h_i^{(L-1)}}} \boldsymbol{\delta h_i^{(L-1)}} \big) \big(1 - \frac{d_r}{(N-1)(d+\beta)} \big).
    \label{equa:gradient_T3'}
\end{gather}
\end{lemma}

\subsection{Combined Derivation and Heuristics}
\label{subsec:combined_derivation}

By combining the approximations of three terms, we get:
\begin{gather}
    \boldsymbol{\delta f_r} \approx \frac{d_r}{d_r+\beta} ||\boldsymbol{(f_r\frac{\partial f_r}{\partial h_r^{(L-1)}}) \circ h_r^{(L-1)}}||_p + k_3 \cdot \delta Topo_r + T_3',
    \label{equa:gradient6}
\end{gather}
where $T_3'$ is Equation~\ref{equa:gradient_T3'}. Here we remove the negative sign before $T_1$, because the original influence is based on the $\ell_1$-norm of the vector of the prediction change, but our approximation of the first and second term only results in a scalar. In reality, $-T_1$ might represent the decrease of the predicted probability of some classes that are related to the removed node, while $T_2$ might represent the increased predicted probability of other classes. To include both of their contributions instead of counteracting them, we directly add the approximated scalar for $T_1$ and $T_2$.

We successfully extend the original formula's form from the L-th layer to (L-1)-th layer by Term $T_3$. By repeating this process, we can approximate $\boldsymbol{\delta f_r}$ by the gradient from every layer. We show the extension process in the appendix. By extending Equation~\ref{equa:gradient6} to all previous layers, we derive:
\begin{small}
\begin{gather}
    \influence(v_r) \approx \sum_{i=0}^{L-1} (\hat{d}_r ^{(L-1-i)}  \hat{h}_r^{(i)}) + k_3' \cdot \delta Topo_r, \nonumber \\
    where\ \hat{d}_r = 1 - \frac{d_r}{(N-1)(d+\beta)},\quad \hat{h}_r^{(i)} = \frac{d_r}{d_r+\beta}
    ||\boldsymbol{(f_r\frac{\partial f_r}{\partial h_r^{(i)}}) \circ h_r^{(i)}}||_p.
    \label{equa:grad_final}
\end{gather}
\end{small}
$\boldsymbol{h_i^{(0)}}$ is the input feature of node $v_i$. We aggregate $\delta Topo_r$ in all layers and change the hyper-parameter $k_3$ to $k_3'$ as a result.

An intuition is that the way we derive Equation~\ref{equa:grad_final} corresponds to the three kinds of node influence in Figure~\ref{fig:model_algorithm_steps}. Approximation of $T_1$ entails node embeddings and gradients, which corresponds to the disappearance of $v_r$'s embeddings. Approximation of $T_2$ contributes to the structural influence $\delta Topo_r$, corresponding to the change of aggregation terms. Approximation of $T_3$ contributes to extending the formula from the last layer to former layers, which corresponds to the influence spreading out, since the neighborhood size grows as the GNN layers increase.

Nonetheless, we cannot compute the approximation for all nodes simultaneously. We change $\boldsymbol{f_r} = \sum_{i=1, i \neq r}^N \boldsymbol{h_i^{(L)}}$ to $\boldsymbol{f} = \sum_{i=1}^N \boldsymbol{h_i^{(L)}}$, so that all the nodes share the same $f$. This will change $\boldsymbol{\frac{\partial f_r}{\partial h_r^{(i)}}}$, but the change might be similar to all nodes due to the following lemma.

\begin{lemma}
If we remove the nonlinear activation function in the GNN layer in Equation~\ref{equa:gradient1}, the gradient $\boldsymbol{\frac{\partial h_r^{(L)}}{\partial h_r^{(L-1)}}}$ can be calculated as:
\begin{gather}
    \boldsymbol{\frac{\partial h_r^{(L)}}{\partial h_r^{(L-1)}}} = \boldsymbol{W_u^{(L)} (W_s^{(L)}} + \alpha_{rr}^{(L)} \boldsymbol{W_m^{(L)}}).
\end{gather}
\end{lemma}

The gradient change caused by adding $\boldsymbol{h_r^{(L)}}$ to $\boldsymbol{f_r}$ might highly rely on the model parameters, so it might be similar among all nodes and does not impact the influence comparison. Though it theoretically does not work for nonlinear GNNs, it could be a useful intuition. Now, we can approximate the influence score for all nodes simultaneously. After one backpropagation, we get $\boldsymbol{\frac{\partial f}{\partial h_r^{(i)}}}$ for every node $v_r \in V$. Then, we can compute the approximation based on Equation~\ref{equa:grad_final}. These calculations are tensor calculations, which only take a few seconds on GPU.


The approximation of node influence on edges in the link prediction task is similar. We replace the sum of predicted node probabilities, $\boldsymbol{f}$, with the sum of link predictions.

\subsection{Complexity Analysis}

We analyze the time and space complexity of the ground truth method (brute-force) and {\model}. $N$ denotes node number, $M$ denotes edge number, $L$ denotes the number of GNN layers, and $h$ denotes the hidden size. In most cases, the adjacency matrix is sparsely stored, and in this situation, according to Blakely et al.~\cite{gnn_complexity}, the time complexity of one forward or backward propagation of a common message-passing GNN model is $O(LNh^2+LMh)$, and the space complexity is $O(M + Lh^2 + LNh)$. We list the time and space complexities in Table~\ref{tab:complexity}. {\model} costs significantly less time than the brute-force method, and basically the same space complexity, so it can be generalized to large real-world graphs.

\begin{table}[t]
    \centering
    \small
    \caption{Complexity comparison.}
    \label{tab:complexity}
    \scalebox{1.0}{
    \begin{tabular}{lrr}
    \toprule
        Method & Time & Space \\  
    \midrule
     Brute-force & $O(LN^2h^2 + LNMh)$ & $O(M + Lh^2 + LNh)$ \\
     {\model} & $O(LNh^2+ LMh)$ & $O(M + Lh^2 + LNh)$ \\
    \bottomrule
    \end{tabular}
    }
\end{table}


\section{Experiments}

\subsection{Baseline Adaption}

There is no mature baseline for this new problem, so we adapt two methods from graph counterfactual explanation as baselines.

\noindent \textbf{Node mask.} Mask optimization is widely used in graph counterfactual explanation~\cite{CF-GNNExplainer,GNNExplainer,CF2,RCExplainer,counterfact_node_edge}. We use a mask vector $\boldsymbol{m} \in \mathbb{R}^N$ to indicate the existence of the $N$ nodes. Elements of $\boldsymbol{m}$ are limited in [0, 1]. In every GNN layer, we multiply node embeddings by $\boldsymbol{m}$ before the message passing. We fix the GNN's parameters and only optimize the mask $\boldsymbol{m}$. Our optimization goal is to maximize the difference between the updated GNN's prediction and its original prediction, calculated as the $\ell_1$-norm. After training, we evaluate the node influence as the distance between elements in $\boldsymbol{m}$ and $\boldsymbol{1}$. Following common designs, we use regularization terms. One regularization term drives the mask elements to zero, which guides elements in $\boldsymbol{m}$ to decrease instead of increase. The second regularization term drives the mask to one, without which the mask might become $\boldsymbol{0}$. Our loss function is:
\begin{equation}
Loss = - \sum_{i=1}^{N} ||\boldsymbol{\gnn(V, E)_i} - \boldsymbol{\gnn(V, E, m)_i}||_1 + \alpha ||\boldsymbol{m}||_1 + \beta ||\boldsymbol{1} - \boldsymbol{m}||_1.
\end{equation}

\noindent \textbf{Prediction model.} A recent work, LARA~\cite{LARA}, greatly improves the counterfactual explanation methods' scalability by applying a GNN model to predict the node/edge influence on the explanation target, so the parameter size is agnostic with the graph size. Inspired by LARA, we train a GCN model to generate a source embedding, $\boldsymbol{p_i}$, and a target embedding, $\boldsymbol{t_i}$ for every node $v_i$. We predict the influence of node $v_i$ on node $v_j$ by $\boldsymbol{p_i} \cdot \boldsymbol{t_j}$, where $\cdot$ is the dot product. We predict the influence of a node as the sum of the predicted influence of its outgoing edges as $\influence(v_r) \approx \sum_{i \in N(r)} \boldsymbol{p_r} \cdot \boldsymbol{t_i}$.

Besides, we also try to directly predict the node influence score by a GNN model, which is a node regression task. In the following tables, ``Predict-E'' is the first way, and ``Predict-N'' is directly predicting node influence. We have tried different GNN models to do the prediction, including GCN and GAT, while GCN performs the best. It might be because we use a few-shot setting. With a very limited number of labels to train the model (7\%), complex GNN structures like GAT might not be well-trained.


\begin{table}[t]
    \centering
    \small
    \caption{Dataset statistics.}
    \label{tab:dataset_statistics}
    \scalebox{0.93}{
    \begin{tabular}{lrrrrrr}
    \toprule
        Dataset & \#Nodes & \#Edges & \#Features & \#Classes & Homo/Hetero \\  
    \midrule
     Cora & 2,708 & 5,429 & 1,433 & 7 & homogeneous \\
     CiteSeer & 3,327 & 4,732 & 3,703 & 6 & homogeneous \\
     PubMed & 19,717 & 44,338 & 500 & 3 & homogeneous \\
     ogbn-arxiv & 169,343 & 1,166,243 & 128 & 40 & homogeneous \\
     P50 & 5,435 & 1,593,721 & one-hot & 2 & heterogeneous \\
     P\_20\_50 & 12,103 & 1,976,985 & one-hot & 2 & heterogeneous \\
    \bottomrule
    \end{tabular}
    }
\end{table}

\begin{table*}[h]
    \centering
    \caption{Pearson correlation coefficient between real influence and approximated influence on citation datasets.}
    \label{tab:citation_result}
    \scalebox{1.0}{
    \begin{tabular}{ll|cccc|cccc}
    \toprule
      & & \multicolumn{4}{c}{Node classification} & \multicolumn{4}{c}{Link prediction} \\
     GNN Model & Method & Cora & CiteSeer & PubMed & ogbn-arxiv & Cora & CiteSeer & PubMed & ogbn-arxiv \\
    \midrule
        \multirow{6}{*}{GCN} 
        & Predict-N & 0.737 & 0.749 & 0.896 & 0.873 & 0.811 & 0.777 & 0.901 & 0.655 \\
        & Predict-E & 0.788 & 0.703 & 0.823 & 0.800 & 0.859 & 0.735 & 0.901 & 0.842 \\
        & Node mask & 0.880 & 0.864 & 0.900 & 0.847 & 0.942 & 0.871 & 0.922 & 0.908 \\
        & {\model-$T_1$} & 0.876 & 0.831 & 0.847 & 0.899 & 0.850 & 0.848 & 0.851 & 0.945 \\
        & {\model-$T_2$} & 0.869 & 0.829 & 0.927 & 0.952 & 0.946 & 0.911 & 0.947 & 0.977 \\
        & {\model} & \textbf{0.903} & \textbf{0.901} & \textbf{0.927} & \textbf{0.956} & \textbf{0.967} & \textbf{0.926} & \textbf{0.949} & \textbf{0.977} \\
    \midrule
        \multirow{6}{*}{GraphSAGE} 
        & Predict-N & 0.712 & 0.709 & 0.808 & 0.856 & 0.693 & 0.595 & 0.877 & 0.52 \\
        & Predict-E & 0.775 & 0.794 & 0.792 & 0.833 & 0.930 & 0.891 & 0.923 & 0.835 \\
        & Node mask & 0.825 & \textbf{0.892} & \textbf{0.896} & 0.878 & 0.816 & 0.305 & 0.948 & 0.734 \\
        & {\model-$T_1$} & 0.829 & 0.819 & 0.816 & 0.943 & 0.944 & 0.903 & 0.813 & 0.898 \\
        & {\model-$T_2$} & 0.859 & 0.838 & 0.831 & 0.956 & 0.923 & 0.842 & 0.933 & 0.927 \\
        & {\model} & \textbf{0.896} & 0.889 & 0.860 & \textbf{0.957} & \textbf{0.978} & \textbf{0.934} & \textbf{0.971} & \textbf{0.936} \\
    \midrule
        \multirow{6}{*}{GAT/DrGAT} 
        & Predict-N & 0.690 & 0.722 & 0.867 & 0.685 & 0.734 & 0.726 & 0.844 & 0.526 \\
        & Predict-E & 0.842 & 0.754 & 0.764 & 0.777 & 0.918 & 0.857 & 0.910 & 0.845 \\
        & Node mask & 0.878 & 0.834 & 0.836 & 0.783 & 0.952 & 0.860 & 0.906 & 0.617 \\
        & {\model-$T_1$} & 0.916 & 0.828 & 0.829 & 0.147 & 0.958 & 0.815 & 0.877 & 0.799 \\
        & {\model-$T_2$} & 0.891 & 0.886 & 0.907 & 0.909 & 0.930 & 0.927 & 0.862 & 0.828 \\
        & {\model} & \textbf{0.927} & \textbf{0.904} & \textbf{0.910} & \textbf{0.909} & \textbf{0.982} & \textbf{0.933} & \textbf{0.951} & \textbf{0.884} \\
    \midrule
        \multirow{6}{*}{GCNII} 
        & Predict-N & 0.729 & 0.739 & 0.824 & 0.873 & 0.794 & 0.777 & 0.882 & 0.718 \\
        & Predict-E & 0.740 & 0.769 & 0.816 & 0.765 & 0.912 & 0.809 & 0.914 & 0.822 \\
        & Node mask & 0.860 & 0.881 & \textbf{0.898} & 0.827 & 0.946 & 0.867 & 0.935 & 0.733 \\
        & {\model-$T_1$} & 0.702 & 0.884 & 0.828 & 0.910 & 0.931 & 0.804 & 0.875 & 0.940 \\
        & {\model-$T_2$} & 0.811 & 0.908 & 0.847 & 0.953 & 0.962 & 0.903 & 0.948 & 0.987 \\
        & {\model} & \textbf{0.874} & \textbf{0.919} & 0.874 & \textbf{0.957} & \textbf{0.969} & \textbf{0.916} & \textbf{0.957} & \textbf{0.987} \\
    \bottomrule
    \end{tabular}
    }
\end{table*}

\begin{table}[h]
    \centering
    \caption{Pearson correlation coefficient between real influence and approximated influence on Twitter datasets.}
    \label{tab:twitter_result}
    \scalebox{0.95}{
    \begin{tabular}{l|cccc}
    \toprule 
        Method & P50 node & P\_20\_50 node & P50 link & P\_20\_50 link \\  
    \midrule
     Predict-N & 0.405 & 0.119 & 0.526 & 0.724 \\
     Predict-E & 0.738 & 0.727 & 0.791 & 0.806 \\
     Node mask & \textbf{0.971} & 0.652 & \textbf{0.968} & \textbf{0.942} \\
     {\model-$T_1$} & 0.951 & 0.751 & 0.910 & 0.903 \\
     {\model-$T_2$} & 0.625 & 0.764 & 0.684 & 0.813 \\
     {\model} & 0.953 & \textbf{0.849} & 0.912 & 0.914 \\
    \bottomrule
    \end{tabular}
    }
\end{table}

\subsection{Experiment Settings}


\textbf{Datasets.}
We use six representative datasets. They include four widely used citation datasets (Cora, CiteSeer, and PubMed~\cite{Cora}, and ogbn-arxiv~\cite{OGB}) and two Twitter datasets (P50 and P\_20\_50~\cite{TIMME}). The four citation networks are homogeneous undirected graphs. One node is a paper, and an edge represents citation. The original task is to predict the research field of each paper (node classification). We add a link prediction task, and we use the dot product of two nodes' representations plus a sigmoid function to do the prediction. The two Twitter datasets are heterogeneous directed graphs. One node is a user, and an edge is one of five Twitter interactions or their counterparts (e.g., be followed): follow, retweet, like, reply, and mention. It has both node classification (predicting user's political leaning) and link prediction, so we use their original tasks. Table~\ref{tab:dataset_statistics} lists the dataset statistics.
An issue is that the trained GNN model is biased to the training-set nodes/edges. To fairly evaluate node influence, we run each experiment 5 times and cycle the data split of nodes/edges by 20\% per time, giving every node/edge an equal chance to show up in training, validation, or test sets. We take the mean of the 5 results.

\noindent \textbf{GNN Models.}
We use six representative GNN models. On the four citation datasets, we use GCN~\cite{GCN}, GraphSAGE~\cite{graphsage}, GAT~\cite{GAT}, and GCNII~\cite{GCNII}. As the ogbn-arxiv dataset is a heated OGB benchmark, we use the SOTA model at the time we started this project, DrGAT~\cite{DrGCN}, to replace GAT in the node classification task. DrGAT is an improved variant of GAT, which has an additional dimensional reweighting mechanism. Since the two Twitter datasets are heterogeneous, the above models can no longer be directly used, so we use TIMME model which is proposed in the same paper as the Twitter datasets~\cite{TIMME}. TIMME tackles three challenges on the Twitter datasets: sparse feature, sparse label, and heterogeneity. We use the original hyper-parameters for DrGAT and TIMME, since they have been carefully tuned. We tune the hyper-parameters for GCN, GraphSAGE, GAT, and GCNII. We also tune the hyper-parameters of each approximation method for each dataset and model.

\noindent \textbf{Evaluation.}
Label generation by the brute-force method is time-consuming, so we evaluate the methods' performance in the few-shot setting, which is more suitable for real-world applications. The methods only have access to 10\% of real influence scores, and the other 90\% are for testing. The ``node mask'' method and {\model} use the 10\% as the validation set to tune hyper-parameters. The ``prediction'' method uses 7\% to train and 3\% for validation. The evaluation metric is the Pearson correlation coefficient between the real influence and the approximated/predicted influence.

\subsection{Performance Comparison}

Table~\ref{tab:citation_result} shows the results on the four citation datasets, and Table~\ref{tab:twitter_result} shows the results on the two Twitter datasets. {\model} outperforms the baseline methods in most cases. Among the baseline methods, the ``node mask'' method performs the best. It is more useful in its original design, which is to analyze the influence on a few nodes or edges. As we want to capture the influence of a node on the whole graph, different nodes/edges are dominantly influenced by different nodes, so it is more difficult to optimize the mask. The ``prediction'' method is greatly limited by label usage. It requires additional ground truth to train, but since we only have a very small training set containing 7\% nodes, its potential is limited. In the appendix, we find that increasing the label usage leads to better performance, but it is still worse than {\model} using 10\% labels.

Besides, we evaluate two variations of {\model} as an ablation study. If we only consider term $T_1$ and term $T_3$ in the approximation (``{\model}-$T_1$'' in the tables), we only consider the task-specific influence of the embeddings but ignore the structural influence, which equals setting $k_3'$ to zero in Equation~\ref{equa:grad_final}. If we only consider term $T_2$ and term $T_3$ (``{\model}-$T_2$'' in the tables), we only consider the structural influence, which equals to setting $k_3'$ to infinity in Equation~\ref{equa:grad_final}. {\model} outperforms both variants, demonstrating the benefit of ensembling the task-specific influence and structural influence. In some situations, only considering one of them could already have a good performance that is close to {\model}. It indicates that sometimes the influence is dominated by the task-specific influence or the structural influence. Ensembling both of them provides us with the opportunity to appropriately utilize both of them.


Here we intuitively analyze the approximation errors of {\model}. The inaccuracy comes from these sources: (1) we only use the first-order derivative for approximation; (2) we use the aggregation term of GCN and GraphSAGE to approximate the aggregation term change. It is inaccurate for more complex aggregation methods like the attention mechanism in GAT;
(3) Equation~\ref{equa:grad_3_2} and Equation~\ref{equa:gradient_T3'} are derived from the rough assumption that every node is functionally and structurally equal, which is not the reality. If the nodes are more diverse, the approximation could be less accurate.
There are also other sources of approximation error, such as using the sum of all output predictions instead of $\boldsymbol{f_r}$, etc. Our analysis could only provide a reference, but the factors are indeed very complex. It is difficult to predict the performance of {\model} given a new dataset or a new GNN model. {\model} contains several inaccurate and intuitive approximations. Nonetheless, its advantage is very high efficiency, and the experiment results on six GNN models and six datasets have already demonstrated its effectiveness.

\subsection{Case Study}

The node classification task on the ogbn-arxiv dataset is to classify each node (paper) into one of forty CS fields defined by the arXiv category (https://arxiv.org/category\_taxonomy). The top-10 influential nodes in the dataset, evaluated by DrGAT model, are listed as below: Adam~\cite{Adam}, ResNet~\cite{residual_learning}, VGGNet~\cite{very_deep_cnn}, an important improvement to Skip-Gram~\cite{improve_skip_gram}, ImageNet~\cite{imagenet}, the paper that proposed word embedding~\cite{word_embedding}, GoogLeNet~\cite{GoogLeNet}, the paper that proposed the batch normalization~\cite{batch_normalization}, the Caffe framework~\cite{caffe}, and Faster R-CNN~\cite{faster_r_cnn}. They are all well-known papers that revolutionized the related fields. It is reasonable that removing them would result in the change of predicted categories of related papers.

        

\subsection{Time Consumption}

\begin{table}[h]
    \centering
    \caption{Time of calculating the real influence.}
    \label{tab:time}
    \scalebox{0.72}{
    \begin{tabular}{l|cccc|cccc}
    \toprule
      & \multicolumn{4}{c}{Node classification} & \multicolumn{4}{c}{Link prediction} \\
      Dataset & GCN & SAGE & (Dr)GAT & GCNII & GCN & SAGE & GAT & GCNII \\
    \midrule
        Cora & 41s & 23s & 42s & 47s & 35s & 25s & 61s & 48s \\
        CiteSeer & 67s & 39s & 46s & 70s & 44s & 35s & 73s & 100s \\
        PubMed & 403s & 382s & $\approx$15min & $\approx$15min & 391s & 425s & $\approx$12min & $\approx$17min \\
        ogbn-arxiv & $\approx$9h & $\approx$9h & $\approx$41h & $\approx$10h & $\approx$2.5h & $\approx$3h & $\approx$13h & $\approx$10h \\
        P50 & \multicolumn{4}{c}{TIMME model: 14min} & \multicolumn{4}{c}{TIMME model: $\approx$1.5h} \\
        P\_20\_50 & \multicolumn{4}{c}{TIMME model: $\approx$41min} & \multicolumn{4}{c}{TIMME model: $\approx$4h} \\
    \bottomrule
    \end{tabular}}
\end{table}

To provide a reference of time consumption, we list the time cost of the brute-force method to calculate the real node influence scores in Table~\ref{tab:time}. The time cost is positively related to the graph size, and the ogbn-arxiv dataset containing 169,343 nodes costs the longest time. It takes about 41 hours for the DrGAT model on the ogbn-arxiv dataset. In comparison, {\model} and the two ``prediction'' baseline methods only need less than one minute. The ``node mask'' baseline method takes less than six minutes. Since we use 10\% of labels as the training and/or validation sets, the total time cost is dominated by generating the ground truth on large graphs.

\subsection{Stability of the Proposed Influence Score}

We want to examine whether this new definition of node influence is stable. We evaluate the stability of the real node influence across different GNNs and different hyper-parameters. We choose an important hyper-parameter, hidden size. We conduct experiments on the node classification task on the four citation datasets. we use GCN~\cite{GCN}, GraphSAGE~\cite{graphsage}, and GAT~\cite{GAT} in this experiment. As in previous experiments, we replace the GAT model with the DrGAT model on ogbn-arxiv. For each model, we use three different hidden sizes: 128, 256, and 512, except for DrGAT on ogbn-arxiv, which only uses 128 and 256 due to memory limitations. For each model and each dataset, we calculate the Pearson correlation coefficient of the influence scores of every pair of different hidden sizes, and we report the mean of those correlation coefficients. These results are in the left three columns in Table~\ref{tab:transfer}. When measuring the cross-GNN stability, for each hidden size and each dataset, we calculate the Pearson correlation coefficient of the influence scores of every pair of two different GNNs, and we report the mean of them in the last column (``Inter-model'') in Table~\ref{tab:transfer}.

\begin{table}[t]
    \centering
    \caption{Stability results.}
    \label{tab:transfer}
    \scalebox{0.95}{
    \begin{tabular}{l|cccc}
    \toprule 
        Dataset & GCN & GraphSAGE & GAT/DrGAT & Inter-model \\  
    \midrule
     Cora & 0.9956 & 0.9857 & 0.9393 & 0.8765 \\
     CiteSeer & 0.9968 & 0.9931 & 0.9585 & 0.8167 \\
     PubMed & 0.9970 & 0.9963 & 0.9451 & 0.8372 \\
     ogbn-arxiv & 0.9984 & 0.9979 & 0.9914 & 0.9557 \\
    \bottomrule
    \end{tabular}
    }
\end{table}

The node influence scores generated by the same GNN with different hidden sizes are quite similar, which demonstrates the stability of the proposed node influence score. The influence scores generated by different GNNs are less similar. In the ideal case, if the GNNs could accurately capture the underlying information spreading patterns of the original graph, then their generated node influence scores should be the same. However, they can not achieve 100\% accuracy. How much the calculated node influence could reveal the actual node influence depends on the accuracy of the GNN model as our surrogate to capture the real information spreading patterns.

\section{Conclusion}

We provide a new perspective of evaluating node influence: the task-specific node influence on GNN model's prediction based on node removal. We use graph neural network (GNN) models as a surrogate to learn the underlying message propagation patterns on a graph. After training a GNN model, we remove a node, apply the trained GNN model, and use the output change to measure the influence of the removed node.
To overcome the low efficiency of the brute-force method (ground truth), we analyze how GNN's prediction changes when a node is removed, decompose it into three terms, and approximate them with gradients and heuristics. The proposed method {\model}
can efficiently approximate node influence for all nodes after only one forward propagation and one backpropagation.
We conduct extensive experiments on six networks and demonstrate {\model}'s effectiveness and efficiency. No matter how we evaluate node influence, we can never touch the real node influence but can only model it, so the modeling perspective is very important. This paper provides a novel perspective of evaluating node influence and offers an intuitive, simple, yet effective solution. Future works are required to better understand node influence and improve the approximation performance.
\begin{acks}
This work was partially supported by NSF 2211557, NSF 1937599, NSF 2119643, NSF 2303037, NSF 20232551, NASA, SRC JUMP 2.0 Center, Cisco research grant, Picsart Gifts, and Snapchat Gifts.
\end{acks}

\bibliographystyle{ACM-Reference-Format}
\bibliography{reference}

\appendix

\section{Supplementary Methods}
\label{sec::appendix_method}

In Section~\ref{subsec:combined_derivation}, we combine the approximation of the three terms to form Equation~\ref{equa:gradient6}, then we extend it to fronter layers and acquire Equation~\ref{equa:grad_final}. Here we explain how to extend the formula to previous layers. In the $T_3'$ in Equation~\ref{equa:gradient6}, we have $\sum_{i \neq r} \boldsymbol{\frac{\partial f_r}{\partial h_i^{(L-1)}} \delta h_i^{(L-1)}}$. Its form is very similar to the form of $\delta f_r$:

\begin{equation}
    \boldsymbol{\delta f_r} = \sum_{i \neq r} \boldsymbol{\delta h_i^{(L)}} = \sum_{i \neq r} \boldsymbol{\frac{\partial f_r}{\partial h_i^{(L)}} \delta h_i^{(L)}}.
    \label{equa:append1}
\end{equation}

We can use the same method of approximating $\sum_{i \neq r} \boldsymbol{\frac{\partial f_r}{\partial h_i^{(L)}} \delta h_i^{(L)}}$ to approximate $T_3'$. After this approximation, we extend the formula to the (L-2)-th layer. By repeating this approximation method layer by layer, we can approximate previous layers similarly and extend the formula to previous layers. When we reach the first GNN layer (the layer after the input), we get:

\begin{small}
\begin{gather}
    \influence(v_r) \approx \sum_{i=0}^{L-1} \big( \hat{d}_r ^{(L-1-i)} (\hat{h}_r^{(i)} + k_3 \cdot \delta Topo_r) \big) + \hat{d}_r ^L \cdot (\sum_{i \neq r} \boldsymbol{\frac{\partial f_r}{\partial h_i^{(0)}} \delta h_i^{(0)}}), \nonumber \\
    where\ \hat{d}_r = 1 - \frac{d_r}{(N-1)(d+\beta)},\quad \hat{h}_r^{(i)} = \frac{d_r}{d_r+\beta}
    ||\boldsymbol{(f_r\frac{\partial f_r}{\partial h_r^{(i)}})} \circ \boldsymbol{h_r^{(i)}}||_p.
    \label{equa:grad_final_2}
\end{gather}
\end{small}



In the formula, $\boldsymbol{h_i^{(0)}}$ is the input feature of $v_i$. It does not change when another node is removed, so $\boldsymbol{\delta h_i^{(0)}} = \boldsymbol{0}$. Besides, since $\delta Topo_r$ is the same in every layer and is only determined by the graph structure, we extract it from the summation, and we re-assign its weight to be $k_3'$. In this way, we can get the final formula of approximating node influence as Equation~\ref{equa:grad_final}.

\section{Supplementary Experiments}

\subsection{Enlarging Label Usage of the ``Prediction'' Method}

As we analyzed previously, the baseline method of predicting node influence by GCN is greatly limited by label scarcity due to our few-shot setting. We explore the influence of enlarging label usage on its performance. We conduct experiments on the (Dr)GAT model on the ogbn-arxiv dataset whose label generation takes the longest time. We use the DrGAT~\cite{DrGCN} model for node classification and the GAT~\cite{GAT} model for link prediction. We keep the ratio of training: validation as 7:3, and we increase the label usage ratio from 10\% to 20\%, 30\%, and 40\%. We list the performance in Table~\ref{tab:gnn_baseline_extend}. In most cases, the model performance increases when there are more labels until the label number has already been sufficient (e.g., 40\%). With the increased performance, the label generation time also increases a lot. In contrast, {\model} does not require training and only acquires a small validation set to tune hyper-parameters, so it is more scalable on large graphs.

\begin{table}[h]
    \centering
    \caption{Performance of the ``prediction'' method with more label usage.}
    \label{tab:gnn_baseline_extend}
    \scalebox{0.85}{
    \begin{tabular}{l|cccc|cccc}
    \toprule
       & \multicolumn{4}{c}{DrGAT (node classification)} & \multicolumn{4}{c}{GAT (link prediction)} \\
      Method & 10\% & 20\% & 30\% & 40\% & 10\% & 20\% & 30\% & 40\% \\
    \midrule
        Predict-N & 0.685 & 0.698 & 0.718 & 0.780 & 0.526 & 0.535 & 0.557 & 0.578 \\
        Predict-E & 0.777 & 0.802 & 0.812 & 0.795 & 0.845 & 0.777 & 0.828 & 0.693 \\
    \bottomrule
    \end{tabular}}
\end{table}

\subsection{Structural Patterns on Twitter Datasets}

\begin{figure}[tbp]
\centering
\includegraphics[width=0.23\textwidth]{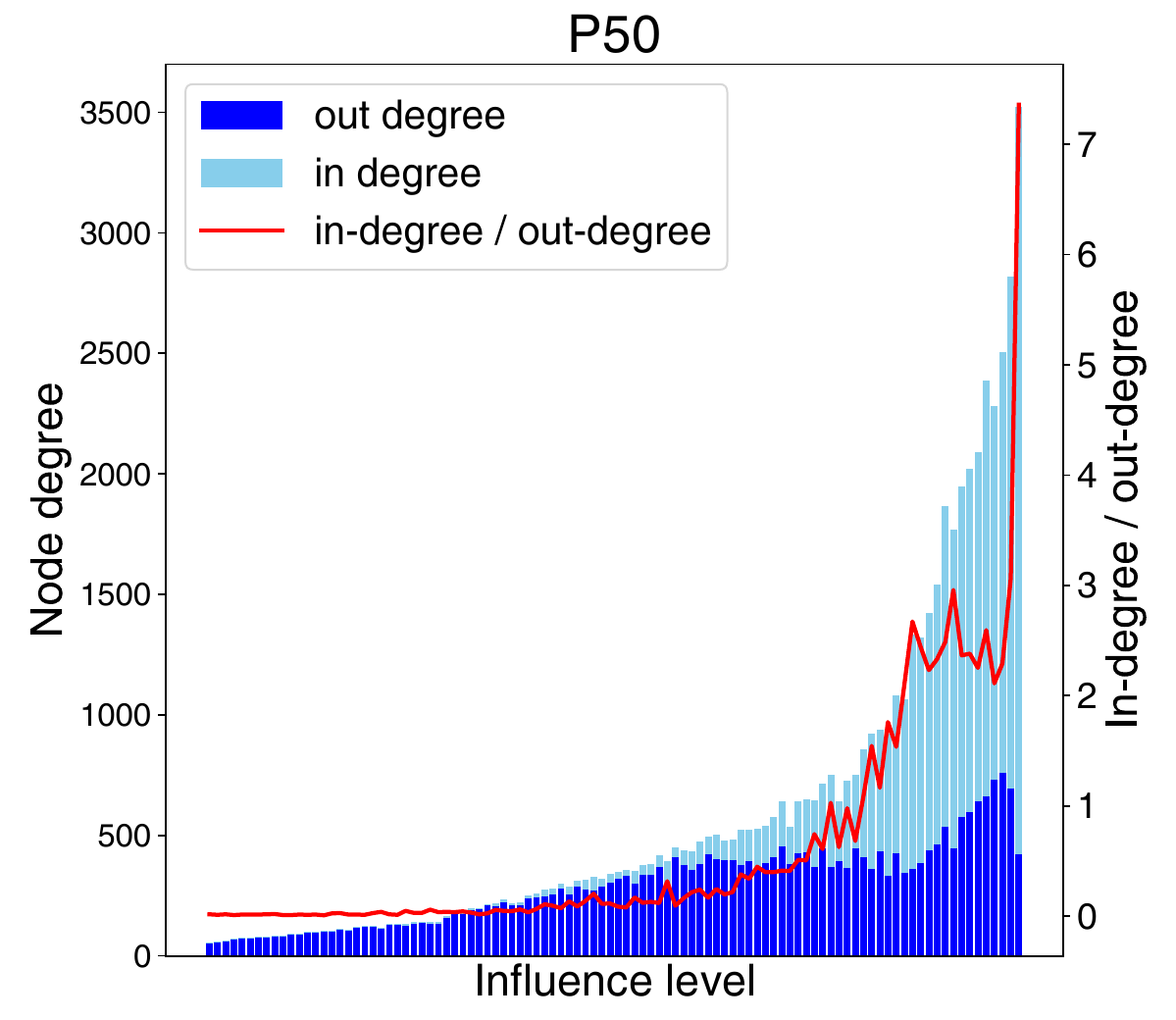}
\includegraphics[width=0.23\textwidth]{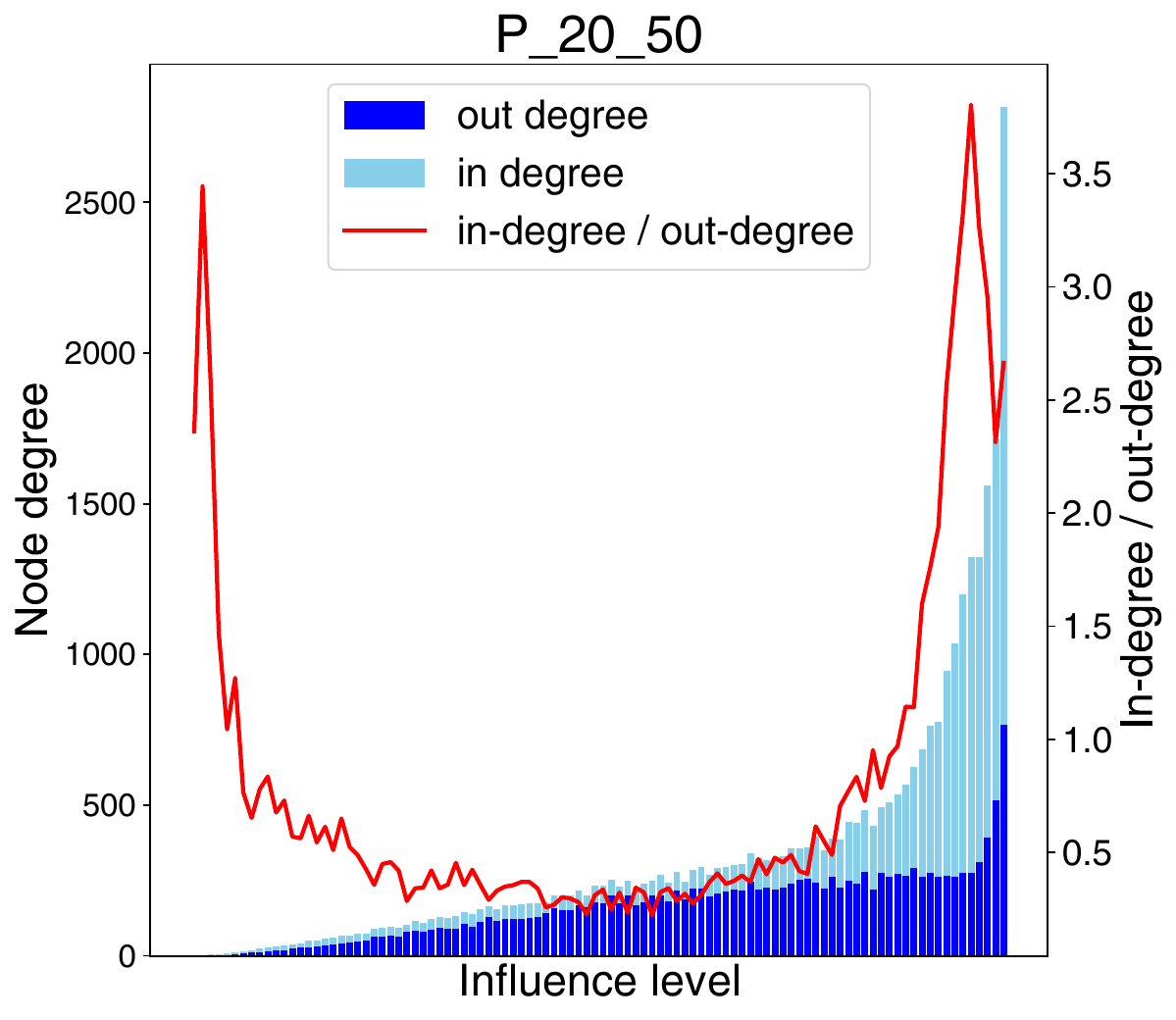}
\caption{Relationship between node influence and degree.}
\label{fig:in_out_degree}
\end{figure}

Since the two Twitter datasets are directed heterogeneous graphs (the four citation datasets are undirected homogeneous graphs), it is worth studying the relationship between structural patterns and node influence on the two Twitter datasets. We first study the relationship between node influence and in-degree or out-degree. For each node, we calculate the mean real influence scores of the two tasks (node classification and link prediction), and we divide the nodes into 100 groups according to their influence level. Then we calculate each group's average in-degree, out-degree, and the ratio of in-degree versus out-degree. We show the results in Figure~\ref{fig:in_out_degree}. Low-influential users have small in-degrees and out-degrees. They might rarely use Twitter. As influence grows, in-degree significantly grows (the light blue area in Figure~\ref{fig:in_out_degree}), while out-degree does not significantly grow from medium-influential to high-influential users (the deep blue area in Figure~\ref{fig:in_out_degree}). It is reasonable since high-influential people get a lot more attention from others than medium-influential people, but high-influential people don't necessarily pay a lot more attention to others. On the P50 dataset, as the influence grows, the ratio of in-degree versus out-degree also significantly grows. However, the ratio of in-degree versus out-degree of the P\_20\_50 dataset is abnormal. The ratio is high in the low-influential groups. It might be due to the randomness when both in-degree and out-degree are very small.

\begin{figure}[htbp]
\centering
\includegraphics[width=0.23\textwidth]{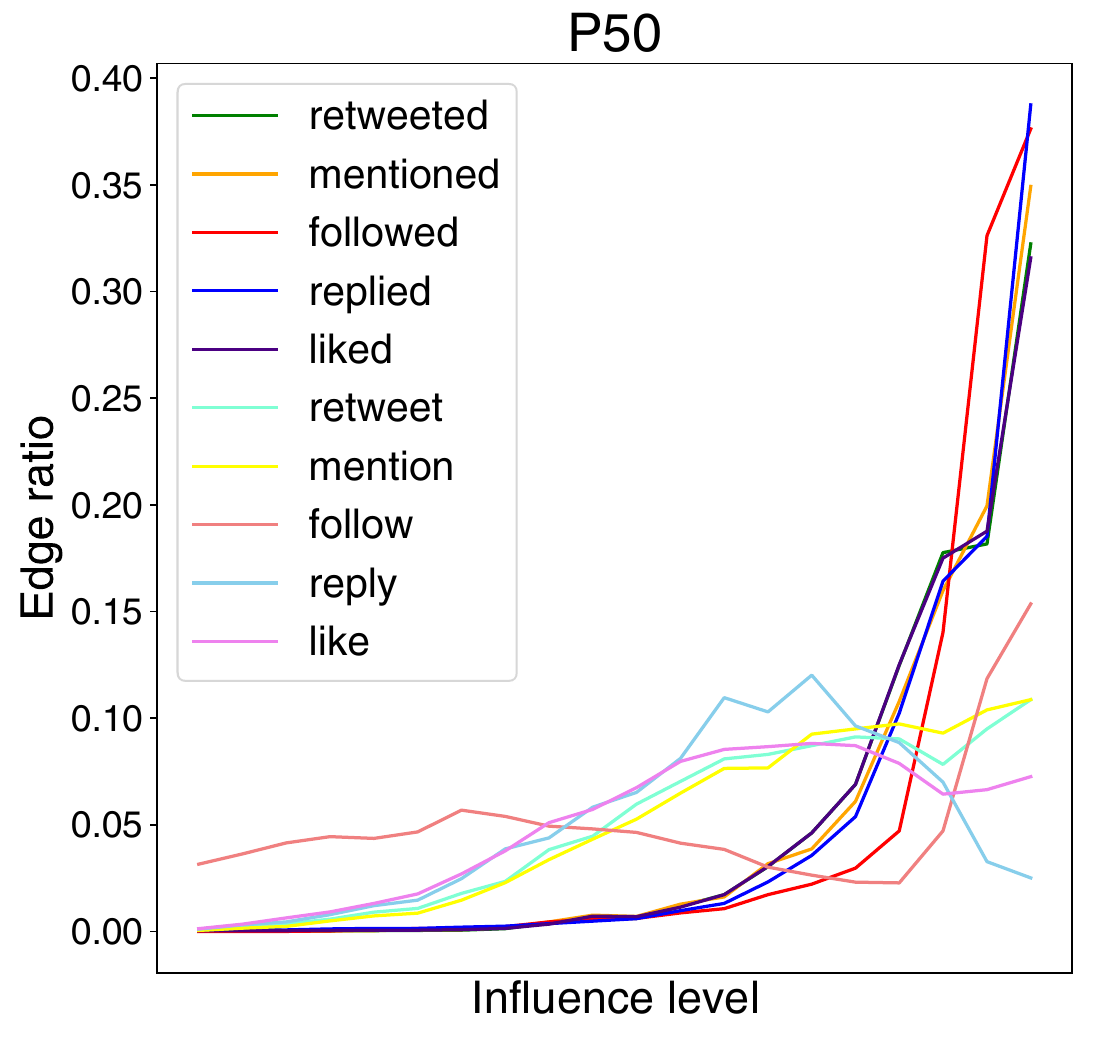}
\includegraphics[width=0.23\textwidth]{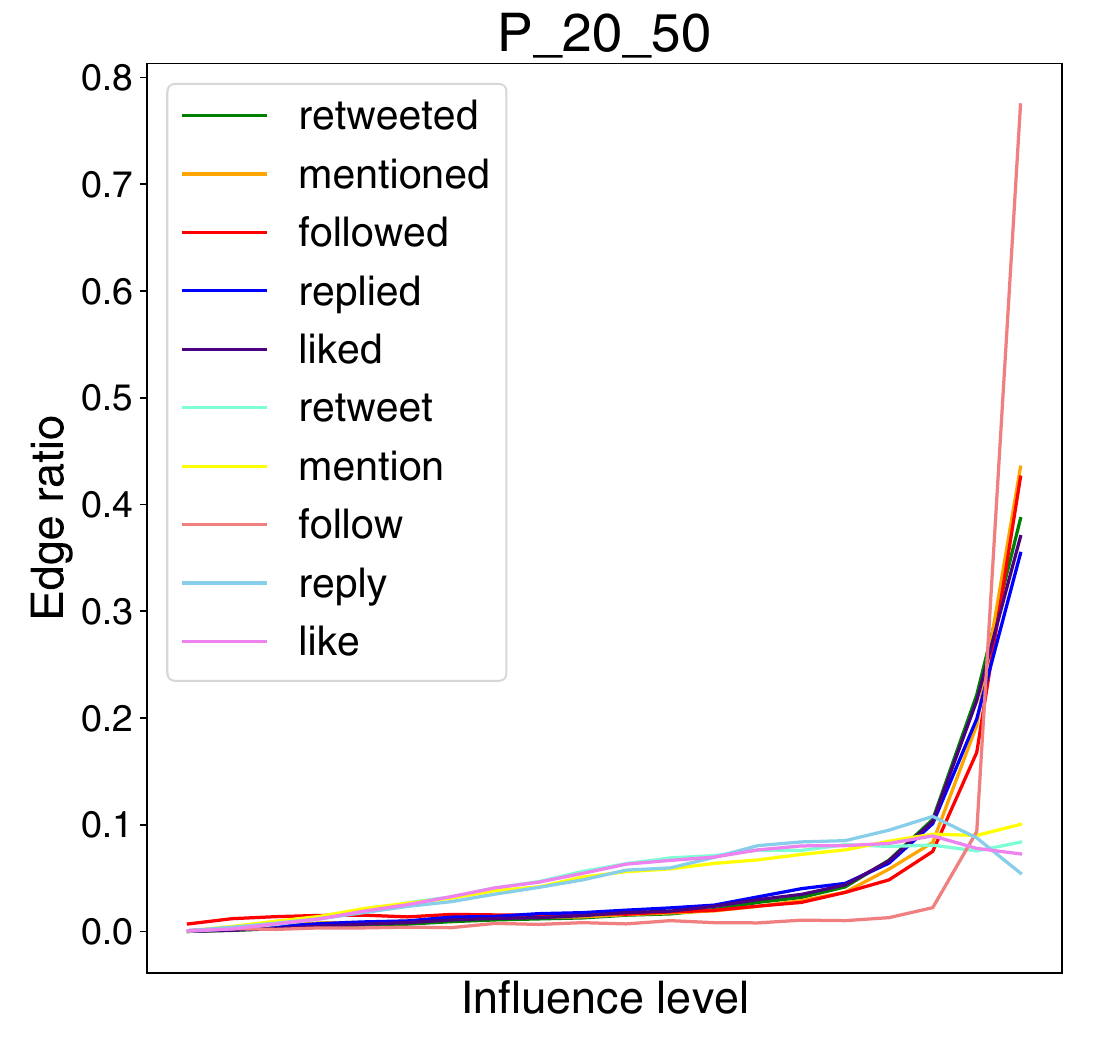}
\caption{The relationship between node influence and edge type. ``Ratio'' means the mean node degree in a certain influence level divided by the total number of edges. Here, the degree and number of edges are separately computed for each edge type.}
\label{fig:relation}
\end{figure}

We also analyze the relationships between node influence and edge type. On the two Twitter datasets, there are 10 types of directional edges: reply, follow, retweet, mention, like, replied by others, followed by others, retweeted by others, mentioned by others, and liked by others. We divide all the nodes into 20 groups according to their influence levels based on the real influence scores. We also use the mean influence scores of the two tasks (node classification task and link prediction task). For every edge type, we calculate the mean degree of nodes within each influence level. Here, when we calculate the node degree for one edge type, we only consider edges of that type. Since different edge types have significantly different numbers of edges, we normalize the node degree by the total number of edges of the corresponding edge type. Since we have separated the ``follow'' relation with ``followed'', ``reply'' with ``replied'', etc., we no longer separately calculate in-degree and out-degree here.

We plot the results in Figure~\ref{fig:relation}. On the P50 dataset, the node degree of ``replied'' and ``followed'' have the strongest positive correlation with the node influence. This observation coincides with TIMME paper~\cite{TIMME}'s observation that ``follow/followed'' and ``reply/replied'' are the two most important relations to predict a user's political leaning. On the P\_20\_50 dataset, ``follow'' replace ``followed'' as the strongest indicator. This might be because the P\_20\_50 dataset is less politics-centered than the P50 dataset. The P50 dataset contains politicians and users who follow or are followed by no less than 50 politicians. The most influential users are probably some politicians, and the number of their followers could be a strong indicator of their influence. In contrast, the P\_20\_50 dataset contains politicians and users who follow or are followed by no less than 20 and less than 50 politicians. The most influential users for GNN's prediction are probably those who follow many politicians and their supporting groups. People usually follow people with the same ideology, so the ``follow'' relationship can be very influential for the TIMME model's prediction, whose node classification target is to predict users' political leanings. Once we remove an account that follows many other people, it might not significantly change other people's political leanings in reality, but the GNN model's predictions might have a big change.




\subsection{Uneven Distribution of Influence}

A real-world application of evaluating task-specific node influence is to use a small number of nodes to trigger a big impact, such as viral advertising~\cite{FirstIM,IM_approximation2,viral_marketing1}, online news dissemination~\cite{news_dissemination1,news_dissemination2}, finding the bottlenecks in an infrastructure network to improve its robustness~\cite{real_removal_example2,real_removal_example3}, etc. In order to achieve these goals, an important feature is that a small number of nodes have a large influence, compared to most nodes having a small influence. Here we analyze the divergence of the proposed node influence. We calculate the ratio of the summed influence of the top k\% influential nodes compared to the sum of all nodes' influence. We calculate the mean results of the real influence scores generated by all GNN models we have used on each dataset. Table~\ref{tab:influence_division} shows the results. A small portion of top influential nodes have a large influence. Node influence on the two Twitter datasets is more unevenly distributed than the four citation datasets. Only 10\% of people contribute to more than 75\% of influence on the node classification task on the Twitter datasets. It is reasonable as a small portion of people on social media attract much more attention than most people. For most datasets, the influence on the node classification task is more unevenly distributed than on the link prediction task. It might be because the prediction of an edge is based on two node embeddings, but the prediction of a node is only based on its own embedding, so node classification might be more sensitive to single-node-removal perturbation than link prediction.

\begin{table}[htbp]
    \centering
    \caption{The ratio (\%) of top k\% influential nodes' summed influence divided by the summed influence of all nodes. ``Node'' represents the node classification task; ``Edge'' represents the link prediction task.}
    \label{tab:influence_division}
    \scalebox{0.85}{
    \begin{tabular}{ll|cccccc}
    \toprule 
        & k\% & Cora & CiteSeer & PubMed & ogbn-arxiv & P50 & P\_20\_50 \\  
    \midrule
     \multirow{3}{*}{Node} &
     1\% & 14.78 & 8.64 & 15.32 & 11.08 & 29.42 & 38.85 \\
     & 3\% & 24.95 & 17.15 & 30.38 & 17.84 & 45.93 & 60.71 \\
     & 10\% & 45.14 & 35.90 & 57.20 & 31.78 & 77.26 & 78.39 \\
     \midrule
     \multirow{3}{*}{Edge} &
     1\% & 12.01 & 8.24 & 10.68 & 19.78 & 21.80 & 24.69 \\
     & 3\% & 19.58 & 15.97 & 21.93 & 29.75 & 34.84 & 42.13 \\
     & 10\% & 34.46 & 32.60 & 45.48 & 46.69 & 60.20 & 63.97 \\
    \bottomrule
    \end{tabular}
    }
\end{table}

\subsection{Experiment Details}

\noindent \textbf{Data Split.} For the node classification task, we use the original data split ratio for ogbn-arxiv, P\_50, and P\_20\_50. For Cora, CiteSeer, and PubMed, in the original data split, the majority of nodes are not in any of the training set, validation set, or test set in the original data split, so we change the data split ratio to train:valid:test = 5:3:2 to cover all nodes.

For the link prediction task, we use the original data split ratio for P50 and p\_20\_50, which is train:valid:test = 8:1:1, and it randomly samples three times negative edges as positive edges. Since there is no original link prediction task on Cora, CiteSeer, PubMed, and ogbn-arxiv datasets, we implement the link prediction task ourselves. We use the same data split ratio of 8:1:1. We only randomly sample the same number of negative edges as positive edges, because we find that too many negative edges on these four datasets would result in model collapse, which simply predicts ``no edge''. It might be because we only implement a simple model, which calculates the dot product of two node embeddings plus a sigmoid function. During training, the model can access the training-set edges, and it needs to predict both the training-set edges and training-set negative edges. During evaluation, the model can access the training-set edges, and it needs to predict the edges and negative edges in the validation set or test set.

\noindent \textbf{Hyper-Parameter Tuning.}
We tune the hyper-parameters for every GNN model and every approximation/prediction method. We use six GNN models: GCN, GraphSAGE, GAT, DrGAT, GCNII, and TIMME. We tune hyper-parameters for GCN, GraphSAGE, GAT, and GCNII. For the DrGAT model and the TIMME model, we use their original hyper-parameters provided by their GitHub repositories, since they have already been carefully tuned by their original papers. Please refer to DrGAT's repository~\footnote{https://github.com/anonymousaabc/DRGCN} and TIMME's repository~\footnote{https://github.com/PatriciaXiao/TIMME} for more details.
We use two-layer GCN, GraphSAGE, GAT, and GCNII when applied to Cora, CiteSeer, and PubMed, and three layers when applied to ogbn-arxiv. On the four citation datasets, the last GNN layer directly provides the output predictions for the node classification task. For the link prediction task, we use the same hidden size in the last layer's output as in hidden layers, and we use the dot product between two nodes' outputs plus a sigmoid function to generate the prediction. The TIMME model contains two GNN layers followed by a node classification MLP or a link prediction module. During hyper-parameter tuning, we mainly tune the hidden size, dropout, learning rate, and weight decay, and we tune the hyper-parameters for every GNN model on every dataset. We store the hyper-parameters in our GitHub repository (https://github.com/weikai-li/NORA.git). Following our notes and running the default codes will automatically use the selected hyper-parameters.

We also tune the hyper-parameters for {\model} and every baseline method, and we tune them for each dataset and each GNN model. {\model} has five hyper-parameters: $k_1$, $k_2$, $k_2'$, $k_3'$, and $\beta$. Among them, $\beta$ is related to approximating term $T_1$ and term $T_3$; $k_1$, $k_2$, and $k_2'$ are related to approximating term $T_2$; $k_3'$ is used to adjust the weight between the two components in Equation~\ref{equa:grad_final}. For convenience, we normalize the two components before multiplying $k_3'$. Based on our experience, we can separately tune the hyper-parameters for the two components, and the best hyper-parameter setting is usually the combination of the best hyper-parameter setting for the two components. $\beta$ is usually within [1, 20]; $k_1$, $k_2$, and $k_2'$ are within [0, 1]; $k_3'$ is usually within [0.5, 5] and not far from 1, since we normalize the two components.

The ``node mask'' baseline method has four hyper-parameters to tune: the learning rate, the two weights $\alpha$ and $\beta$ of its two regularization terms, and the number of training epochs. We usually optimize the mask for about 100 to 300 epochs, and we use the mask vector which achieves the lowest validation loss (not including the regularization loss).

The ``prediction'' baseline method which trains a GCN model to predict node influence has six hyper-parameters to tune: the learning rate, weight decay, number of training epochs, hidden size, number of GCN layers, and the dropout. We usually train the model for 100 to 300 epochs, and we use the model that has the lowest validation loss during training. We use the MSE loss function for this regression task.

We store all the hyper-parameters in our GitHub repository (https://github.com/weikai-li/NORA.git), and we provide scripts for every approximation method for every GNN model and every dataset. With the scripts containing the hyper-parameters, it is easy to reproduce our results.

\end{document}